\documentclass[sigconf]{acmart}
\AtBeginDocument{%
  }

\setcopyright{acmlicensed}
\copyrightyear{2018}
\acmYear{2018}
\acmDOI{XXXXXXX.XXXXXXX}
\acmConference[]{}{}

\usepackage{enumitem}
\setlist[itemize]{leftmargin=10pt, itemsep=1pt}
\usepackage{booktabs}
\usepackage{multirow}
\usepackage[table]{xcolor}

\renewcommand\footnotetextcopyrightpermission[1]{}
\begin{document}

\title{PrefReward: Learning User Preference Matrix for Personalized Text Generation}


\author{Yue Wu}
\email{wuyuewy@mail.ustc.edu.cn}
\affiliation{%
  \institution{University of Science and Technology of China}
  \country{China}
}

\author{Chengbing Wang}
\email{wwq197297@mail.ustc.edu.cn}
\affiliation{%
  \institution{University of Science and Technology of China}
  \country{China}
}

\author{Yimeng Bai}
\email{baiyimeng@mail.ustc.edu.cn}
\affiliation{%
  \institution{University of Science and Technology of China}
  \country{China}
}

\author{Xiaoyan Zhao}
\email{xzhao@se.cuhk.edu.hk}
\affiliation{%
  \institution{The Chinese University of Hong Kong}
  \country{China}
}

\author{Yang Zhang}
\email{zyang1580@gmail.com}
\affiliation{%
  \institution{National University of Singapore}
  \country{Singapore}
}

\author{Fuli Feng}
\email{fulifeng93@gmail.com}
\affiliation{%
  \institution{University of Science and Technology of China}
  \country{China}
}





\renewcommand{\shortauthors}{Wu et al.}

\begin{abstract}
Large Language Models (LLMs) have demonstrated remarkable ability in generating personalized content by leveraging user histories and contextual cues. However, most existing personalization approaches rely on implicit representations within model parameters, making it difficult to interpret user-specific preferences or effectively handle long-context dependencies. To address these challenges, we propose \textbf{PrefReward}, a novel preference-aware generative framework that explicitly models user styles through a structured preference matrix and integrates it into the decoding process as a reward signal. PrefReward consists of two stages: (1) extracting a user-specific preference matrix that summarizes individual stylistic tendencies, and (2) using the matrix to guide generation via a KL-divergence-based reward function. Experiments on the LongLaMP dataset show that PrefReward outperforms retrieval-based baselines and context steering method in both generation quality and personalization interpretability.
\end{abstract}





\maketitle

\section{Introduction}
Large Language Models (LLMs) have demonstrated impressive capabilities in generating fluent, coherent, and context-aware text across a wide range of natural language generation tasks~\cite{LaMP_nlp1,Survey}. Recently, there has been growing interest in applying LLMs to \textbf{personalized text generation}, where the goal is to adapt the generation process to individual user preferences, historical behaviors, or linguistic styles~\cite{personlization1,personlization2,personlization3}. Such personalization is crucial for applications like dialogue systems, writing assistants, and recommendation-based content generation, where user satisfaction depends heavily on stylistic alignment and contextual relevance~\cite{import_long_context,drift-word-list,CoS,RAG_limit_and_CoAS}.


However, LLM personalization presents several challenges. 
\textbf{First}, many existing methods require \textit{additional training} to reliably align model outputs with user preferences — this includes supervised fine-tuning, preference-conditioned adapters, or reinforcement learning procedures. Such training demands labeled preference data, per-user or per-group updates, and substantial compute and engineering effort, which undermines scalability and rapid deployment.  
\textbf{Second}, current personalization techniques suffer from \textit{limited interpretability}. Whether preferences are embedded in aligned weights, latent attention over retrieved contexts, or implicitly induced by prompt engineering, the resulting preference signal is typically opaque: 
it is difficult to inspect which stylistic dimensions the model has learned
to provide human-understandable explanations and controls. 
For example, a writing assistant that preserves a user's preference for concise or expressive wording across tasks can provide more consistent and useful assistance. Explicitly representing these dimensions enables consistent, controllable adaptation across repeated interactions, where a generic response style may not suit every user.

To address these challenges, we propose a new paradigm for personalized generation called \textbf{PrefReward}, which explicitly models and leverages user preferences to guide the decoding process. 
By introducing an explicit preference representation and a quantitative alignment reward, PrefReward bridges the gap between implicit personalization and controllable stylistic generation. It offers two key advantages:  
(1) \textbf{Interpretability}: user preferences are represented as visible, human-understandable matrices rather than opaque latent vectors;  
(2) \textbf{Train-free personalization}: PrefReward operates entirely at the decoding stage without additional fine-tuning or reinforcement learning, making it lightweight and easily applicable to any pretrained LLM backbone.

Our main contributions are summarized as follows:
\begin{itemize}
    \item We propose \textbf{PrefReward}, a novel reward-guided personalization framework that explicitly models user preferences as interpretable matrices and aligns generation with them through KL-divergence optimization.
    \item We design an efficient preference extraction pipeline combining a BM25-based retriever with token-level LLM inference to capture fine-grained stylistic distributions from user histories.
    \item We conduct extensive experiments on the \textbf{LongLaMP} dataset using \textbf{Llama2-7B-Chat} and the aligned \textbf{Gemma-2B-IT}, demonstrating that PrefReward achieves consistent improvements over retrieval-based baselines and context steering method.
\end{itemize}

\section{Problem Formulation}

Let $\mathcal{U}$ denote the set of users. Each user $u \in \mathcal{U}$ is associated with a history profile $\mathcal{H}_u$ containing multiple text samples, representing historical input-output pairs. 
Given a user $u$ and an input prompt $x$ describing a specific task (e.g., in LongLaMP, each record contains a prompt with product context and the user's history review), the language model $\mathcal{M}$ aims to generate an output: 
\begin{equation}
\hat{y} = \mathcal{M}(x, \mathcal{H}_u),
\end{equation}
conditioned on both the prompt $x$ and the user profile: 
\begin{equation}
\mathcal{H}_u = \{(x_i, y_i)\}_{i=1}^{K}.
\end{equation}
Here, each pair $(x_i, y_i)$ encodes a past interaction or text example of the user. The reference output $y$ represents the desired personalized response, reflecting the unique stylistic patterns of user $u$ (e.g., customized product reviews). The goal is to ensure that the generated output $\hat{y}$ not only satisfies the task requirements but also aligns closely with the user-specific writing style inferred from $\mathcal{H}_u$.


\begin{figure}[t]
\centering
\includegraphics[width=1.0\linewidth]{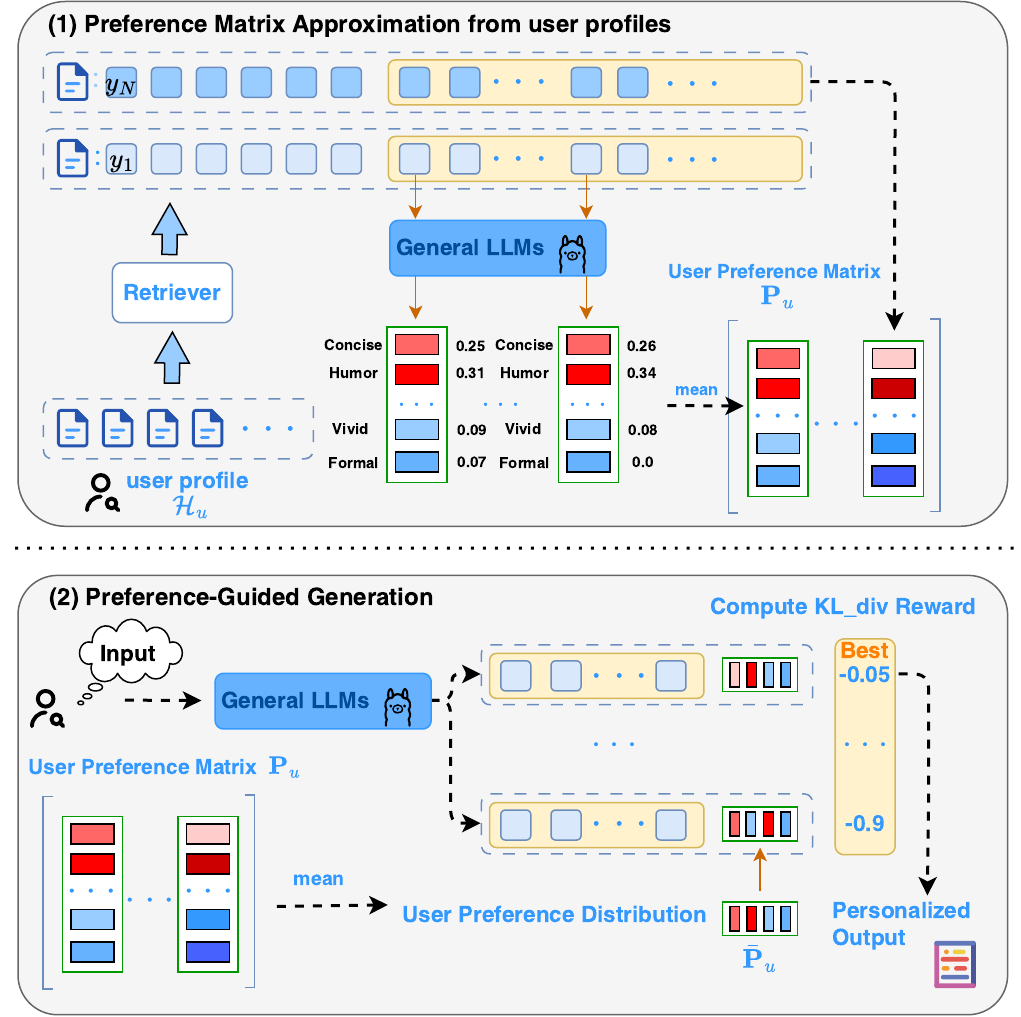}
\caption{Overview of the proposed PrefReward framework, which consists of (1) Preference Matrix Approximation from user profiles and (2) Preference-Guided Decoding using KL-divergence reward.}
\label{fig:framework}
\end{figure}

\section{PrefReward}
 In this section, we present our PrefReward framework for personalized text generation as shown in Figure~\ref{fig:framework} .

\subsection{Preference Matrix Extraction}
\noindent
\textbf{Preference Label Selection}
A central component of our framework is the predefined preference label set $\mathcal{P}$, which determines the stylistic dimensions captured by the user preference matrix.  
To construct a label set that is both expressive and interpretable, we take inspiration from the \textit{Drift} framework~\cite{drift-word-list}, which advocates defining stylistic attributes as explicit, human-understandable categories instead of relying on latent embedding clusters. Our goal is to build a structured representation $\mathbf{P}_u \in \mathbb{R}^{K \times N}$, where $N$ denotes the number of predefined preference labels from a fixed set $\mathcal{P} = \{\textit{formal}, \textit{concise}, \textit{vivid}, \textit{humor}, \ldots\}$.
Drift motivates this broad candidate pool as increasing the likelihood of capturing nuanced user preferences~\cite{drift-word-list}. Accordingly, retaining this diverse attributes provides a richer basis for representing individual stylistic differences and supports fine-grained personalization.

\noindent
\textbf{Retriever for Profile Selection.}
Directly using all historical profiles can introduce noise and redundancy, especially for users with long interaction histories.  
To ensure that only representative and stylistically informative samples are used for preference modeling, we employ a \textbf{retriever module} based on the \textit{BM25} ranking algorithm to select the top-$K$ relevant profiles from $\mathcal{H}_u$.  
Profiles with the highest scores are selected to form $\mathcal{H}^*_u = \{(x_i, y_i)\}_{i=1}^{K}$, which are then used for preference extraction.

\noindent
\textbf{Token-Level Preference Extraction.} For each text sample, we perform token-wise inference with a pre-trained LLM. At step $i$, we feed $(x,y^{<i})$ to the model and restrict the next-token logits to tokens corresponding to labels in $\mathcal{P}$ and obtain:
\begin{equation}
p_\theta(t_i\mid x,y^{<i})=\operatorname{softmax}_{t\in\mathcal{V}_{\mathcal{P}}}(\operatorname{LLM}(x,y^{<i})),
\end{equation}
where $\mathcal{V}_{\mathcal{P}}$ is the set of tokens associated with labels in $\mathcal{P}$. The resulting distribution forms a preference vector $v_i\in\mathbb{R}^{N}$. Averaging these vectors across all tokens yields a sequence-level preference vector $p_y$:
\begin{equation}
\mathbf{p}_y = \frac{1}{|y|}\sum_{i=1}^{|y|} \mathbf{v}_i.
\end{equation}
Aggregating the preference vectors across $K$ samples produces the user preference matrix $\mathbf{P}_u = [\mathbf{p}_{y_1}; \ldots; \mathbf{p}_{y_K}]$.

\noindent
\textbf{Interpretability and Efficiency.}
This matrix encodes interpretable stylistic preferences—each dimension corresponds to an explicit style or tone. Unlike latent embeddings, $\mathbf{P}_u$ can be directly visualized or compared across users. Computation is efficient since token-level logits are already available during forward passes.

\subsection{Preference-Guided Generation}

\textbf{KL-Divergence Reward.}
Given a user $u$ and their preference matrix $\mathbf{P}_u$, we generate $N_c$ candidate outputs $\{y^{(1)}, \ldots, y^{(N_c)}\}$ for an input $x$.  
Each candidate’s preference vector $\mathbf{p}_{y^{(i)}}$ is derived using the same token-level extraction method.  
To compute the alignment reward, we first obtain the user's mean preference distribution:
\begin{equation}
\bar{\mathbf{P}}_u = \frac{1}{K} \sum_{j=1}^{K} \mathbf{p}_{y_j}.
\end{equation}
The reward for each candidate is then computed as:
\begin{equation}
R_i = -\mathrm{KL}\!\left(\mathbf{p}_{y^{(i)}} \,\big\|\, \bar{\mathbf{P}}_u\right).
\end{equation}
The best candidate is selected as $y^* = \arg\max_i R_i$.  
This reward formulation effectively aligns the generation style with user-specific preferences without retraining the base model.

\noindent
\textbf{Decoding with Best-of-N Sampling.}
We employ a Best-of-$N_c$ strategy: for each prompt, the LLM produces $N_c$ stochastic generations, and the candidate with the highest reward is chosen. This strategy balances computational cost and personalization quality.

\section{Related Work}

Inspired by the personalization potential of large language models (LLMs), recent studies have explored incorporating user context into generation. Our work is closely related to retrieval- and decoding-based personalization approaches~\cite{RAG_limit_and_CoAS,align,RAG_Personlization,CoS,drift-word-list}. Retrieval-Augmented Generation (RAG)~~\cite{RAG_Personlization} retrieves relevant user profiles via either BM25~\cite{BM25} or Contriever~\cite{contriver} matching to provide contextual grounding. However, such retrieval-based personalization often struggles in long-context scenarios where retrieved texts fail to capture consistent user style. Context Steering (CoS)~~\cite{CoS} introduces a simple, training-free decoding strategy that amplifies contextual influence by comparing the output probabilities from two forward passes—with and without user context—and linearly scaling their difference. While CoS flexibly controls the degree of personalization, it relies on shallow contextual similarity and lacks explicit preference modeling. Unlike these methods, our proposed \textbf{PrefReward} explicitly constructs user preference matrices and applies KL-divergence-based reward optimization, achieving interpretable and stable personalization across long user histories.

\section{Experiments}

\subsection{Experimental Setup}

\subsubsection{Datasets.}
We conduct experiments on the \textbf{LongLaMP} dataset, which provides user-level profiles and historical interaction records with long-context text samples~\cite{longlamp}. We focus on the personalized review writing task. For each user, We retrieve $K=10$ representative profiles per user for preference extraction.  

\subsubsection{Baselines.}
We compare our proposed \textbf{PrefReward} with several representative baselines:  
\begin{itemize}
    \item \textbf{NonPers}: Standard LLM generation without any personalization or retrieval augmentation. This baseline evaluates the inherent generative ability of the model.
    \item \textbf{RAG (BM25)}: Retrieval-Augmented Generation using BM25 retriever to fetch relevant user profiles for contextual grounding. It provides a retrieval-based personalization reference~\cite{RAG_Personlization}.
    \item \textbf{RAG (CT)}: A retrieval-enhanced method that leverages dense embeddings from the Contriever model to retrieve user history semantically~\cite{contriver}.
    \item \textbf{CoS ($\lambda=1.0, 2.0$)}: Context Steering (CoS) is a training-free decoding method that amplifies user-specific context in generation. It compares outputs with and without context and uses $\lambda$ to control the degree of personalization~\cite{CoS}.
\end{itemize}

\subsubsection{Evaluation Metrics.}
Following prior works~\cite{longlamp} on personalized generation and retrieval-augmented LLMs, we adopt three widely used evaluation metrics to assess generation quality and personalization effectiveness: ROUGE-1, ROUGE-L, and METEOR.

\subsubsection{Implementation Details.}
For zero-shot model experiments, we leverage \textbf{Llama2-7B-Chat} as the backbone model. For aligned model experiments, we align \textbf{Gemma-2B-IT} using a dataset of (prompt, target output) pairs.
These complementary settings test PrefReward both without additional task-specific alignment and on a backbone aligned with task examples.
During preference extraction, we set the preference label list $\mathcal{P}$ to include 41 stylistic dimensions (e.g., \textit{formal}, \textit{concise}, \textit{vivid}, \textit{humor}, \textit{modest}, \textit{energy}, etc.) following Drift's setting~\cite{drift-word-list}.  
For reward-guided decoding, we generate $N_c=10$ candidate outputs per prompt using nucleus sampling ($p=0.95$) and temperature $\tau=0.8$.  

\begin{table}[t]
\centering
\caption{Overall performance comparison on the LongLaMP dataset across two backbone LLMs. Best results under each model are highlighted in bold.}
\label{tab:main_models_vertical}
\resizebox{!}{0.30\linewidth}{
\begin{tabular}{llccc}
\toprule
\textbf{Backbone} & \textbf{Method} & \textbf{ROUGE-1} & \textbf{ROUGE-L} & \textbf{METEOR} \\
\midrule
\multirow{6}{*}{\textit{Zero-shot}} 
  & NonPers           & 0.3139 & 0.1445 & 0.1934 \\
  & RAG (BM25)        & 0.2980 & 0.1417 & 0.1850 \\
  & RAG (CT)  & 0.3022 & 0.1429 & 0.1880 \\
  & CoS ($\lambda$=1.0) & 0.2539 & 0.1245 & 0.1602 \\
  & CoS ($\lambda$=2.0) & 0.2365 & 0.1163 & 0.1513 \\
  \rowcolor[HTML]{F2F2F2}
  & \textbf{PrefReward} & \textbf{0.3161} & \textbf{0.1450} & \textbf{0.1972} \\
\midrule
\multirow{6}{*}{\textit{Aligned}} 
  & NonPers           & 0.2603 & 0.1359 & 0.1431 \\
  & RAG (BM25)        & 0.3276 & 0.1736 & 0.1998 \\
  & RAG (CT)  & 0.3277 & 0.1733 & 0.1999 \\
  & CoS ($\lambda$=1.0) & 0.2869 & 0.1393 & 0.1945 \\
  & CoS ($\lambda$=2.0) & 0.2491 & 0.1193 & 0.1713 \\
  \rowcolor[HTML]{F2F2F2}
  & \textbf{PrefReward} & \textbf{0.3387} & \textbf{0.1751} & \textbf{0.2090} \\
\bottomrule
\end{tabular}
}
\end{table}

\subsection{Overall Results}
Table~\ref{tab:main_models_vertical} reports the overall performance of all methods on both 
\textbf{Llama2-7B-Chat} and the aligned \textbf{Gemma-2B-IT} backbone.  
The following observations can be made:

\begin{itemize}
    \item \textbf{PrefReward consistently achieves the best performance across both models.}  
    On all three metrics, PrefReward surpasses strong personalization baselines, demonstrating that explicit preference matrices combined with KL-divergence reward guidance lead to more faithful and stylistically aligned generation.  
    The gains are stable across backbones, indicating that the proposed method is model-agnostic and generalizes well to different LLM architectures.

    \item \textbf{RAG and CoS remain limited in long-context personalization.}  
    In the zero-shot setting, retrieval-based approaches (BM25 and Contriever) are not effective: retrieval noise limits the usefulness of retrieved histories, and these approaches struggle to capture persistent user-level stylistic tendencies when applied to backbones without task-specific alignment.
    CoS, although training-free and lightweight, relies solely on contextual perturbation and thus struggles to maintain stable personalization when user histories grow longer or more complicated.
\end{itemize}

\noindent
Overall, these findings confirm that explicitly modeling user preferences as decoding rewards provide a more effective and interpretable solution for long-context personalized generation.

\subsection{Sensitivity to Preference Labels Choice}
\begin{figure}[H]
\centering
\includegraphics[width=\linewidth]{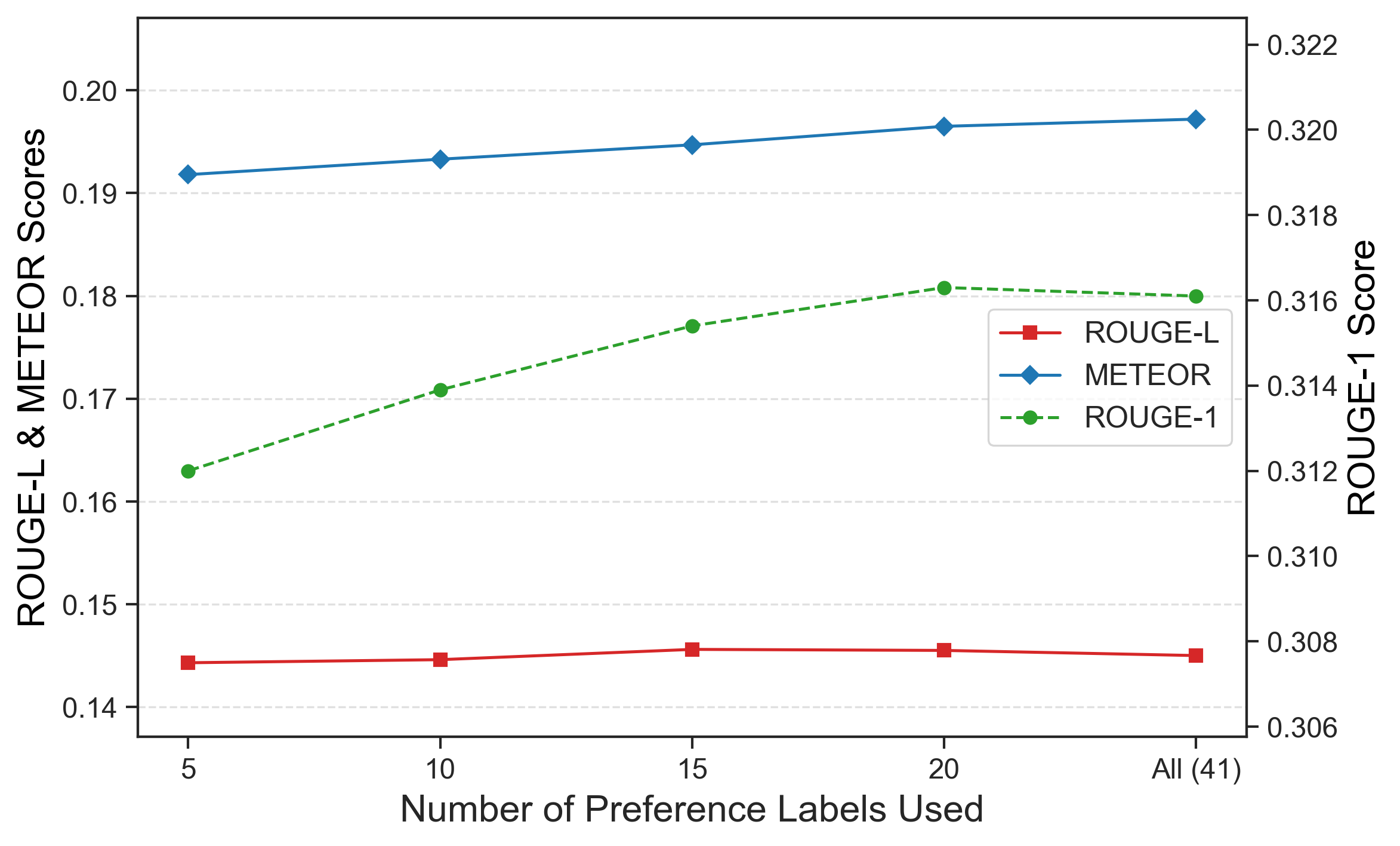}
\caption{Effect of different preference label choice on model performance}
\label{fig:word}
\end{figure}

To investigate how the different choices of preference labels influences model performance,  
we vary the granularity and coverage of user preference categories used to construct the preference matrix,  
Figure~\ref{fig:word} reports the corresponding results on ROUGE-1, ROUGE-L, and METEOR.
These results demonstrate that more comprehensive and finer-grained preference-label sets, such as the full set of 41 labels, better support personalization. By contrast, coarser sets containing only 5 or 10 labels yield weaker personalization, as they may not capture enough distinctions among users' preferences.

\subsection{Overhead analysis}
 
\begin{figure}[t]
\centering
\includegraphics[width=0.95\linewidth]{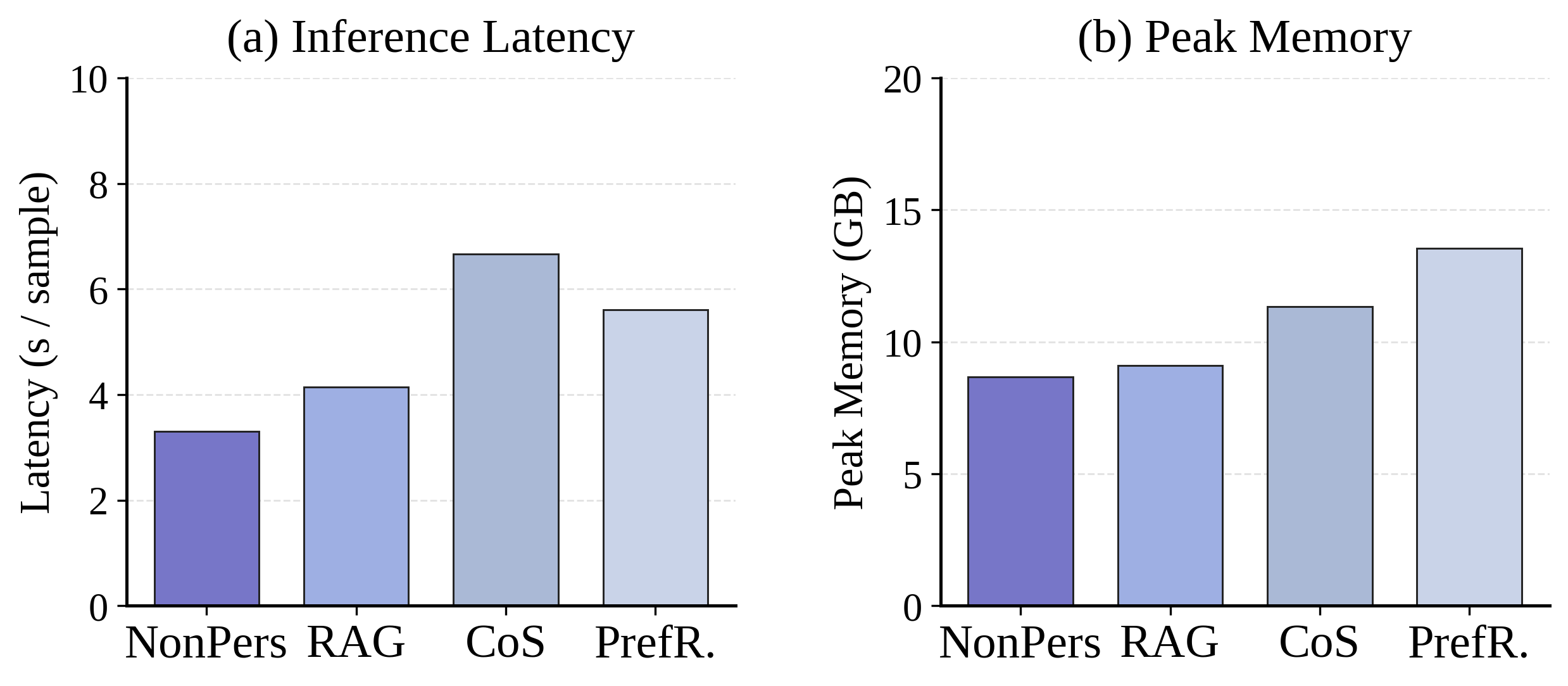}
\caption{Efficiency analysis of PrefReward(PrefR.).}
\label{fig:efficiency}
\end{figure}

\label{sec:efficiency}
We finally assess the computational overhead of PrefReward relative to
RAG and CoS method on Gemma-2B-IT model during inference time. Figure~\ref{fig:efficiency} 
reports the inference latency and peak GPU memory measured on different 
settings. PrefReward
achieves stronger personalization with a bounded and predictable additional scoring cost.
User preference profiles can be extracted and maintained offline, so preference extraction adds no online inference overhead.

\subsection{Case Study}

\begin{figure}[h]
\centering
\includegraphics[width=\linewidth]{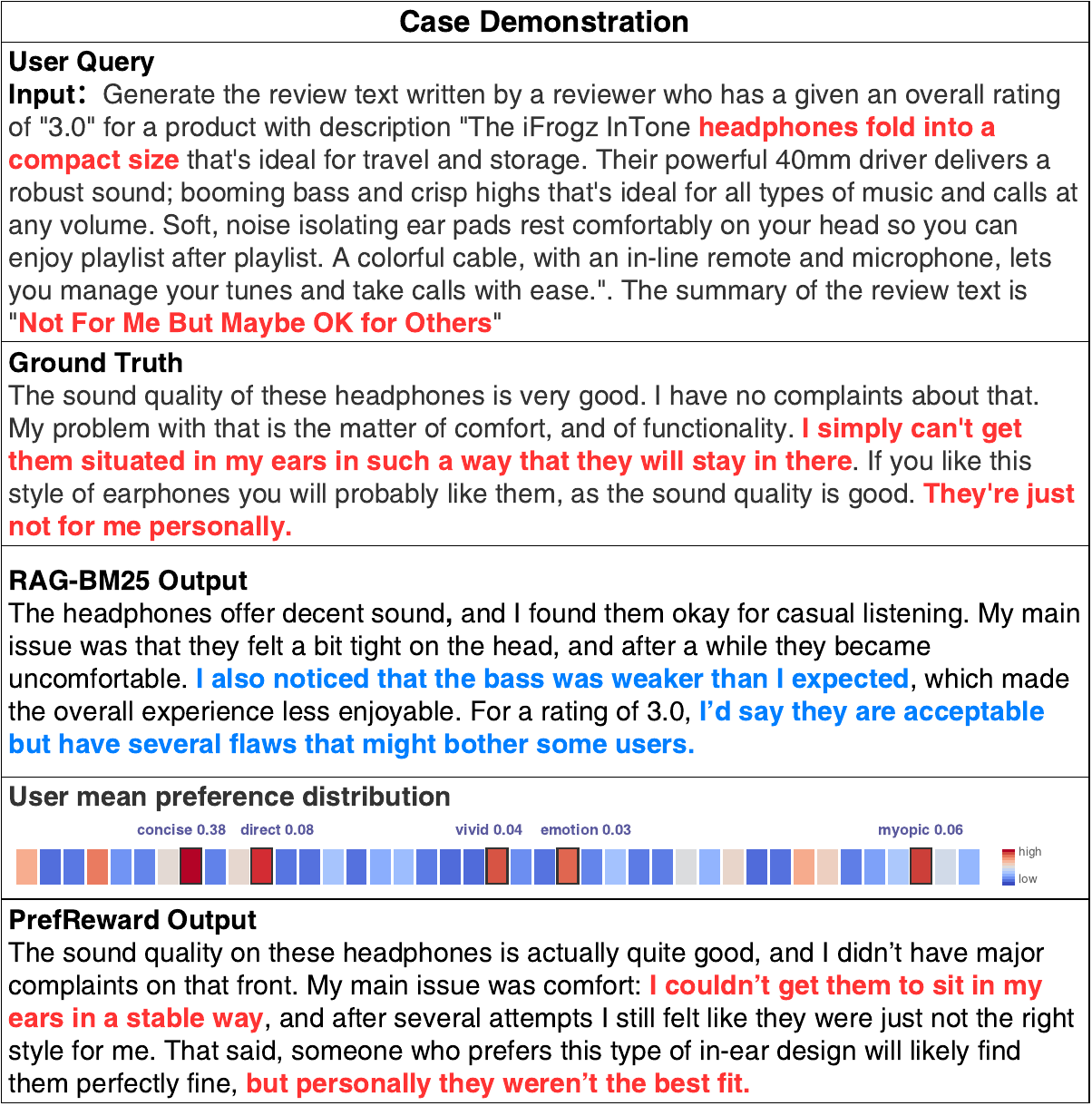}
\caption{Case study of personalized generation in LongLaMP}
\label{fig:case}
\end{figure}

Figure~\ref{fig:case} presents qualitative examples comparing the outputs of baseline models and our PrefReward framework. The PrefReward output closely aligns with the ground-truth review, accurately capturing key user sentiments,
demonstrating stronger controllability and interpretability in personalized generation.

\section{Conclusion}
We introduced \textbf{PrefReward}, a preference-aware framework for personalized text generation. By explicitly modeling user preferences as interpretable matrices and integrating them via KL-divergence rewards, PrefReward bridges the gap between personalization quality and transparency. Experiments confirm its superiority over retrieval-based and non-personalized baselines. Future work will explore fine-tuning LLMs with preference-aware reinforcement learning for end-to-end optimization.

\section{Limitations}

Our PrefReward follows the Drift's predefined preference label setting to capture diverse
user preferences.
The label-choice experiment provides a sensitivity check, while different vocabularies and finer granularity remain possible extensions.
Our evaluation covers
two backbones of up to 7B parameters on LongLaMP benchmark; scaling to
larger models, more benchmarks remains future
work.

\bibliographystyle{ACM-Reference-Format}
\bibliography{sample-base}

\appendix

\end{document}